\documentclass[11pt]{article}

\usepackage[preprint]{acl}

\usepackage{times}
\usepackage{latexsym}

\usepackage[T1]{fontenc}

\usepackage[utf8]{inputenc}

\usepackage{microtype}

\usepackage{inconsolata}

\usepackage{graphicx}
\usepackage{booktabs}

\usepackage{array}
\newcolumntype{L}[1]{>{\raggedright\arraybackslash}m{#1}}
\newcolumntype{C}[1]{>{\centering\arraybackslash}m{#1}}

\usepackage{pgfplots}
\pgfplotsset{compat=1.18}

\title{Behind the \texttt{[MASK]}: Disentangling Representation and Faithfulness in DAPF-Based Dementia Detection}

\author{Pardis Ranjbar-Noiey \\
  University of Illinois Chicago \\
  Chicago, IL, USA  \\
  \texttt{pranjb3@uic.edu} \\\And
  Natalie Parde \\
  University of Illinois Chicago \\
  Chicago, IL, USA \\
  \texttt{parde@uic.edu} \\}

\begin{document}
\maketitle

\begin{abstract}
Spoken-language analysis via prompt-based domain-adaptive models is a promising direction for low-resource, non-invasive dementia screening, but such models remain internally opaque. We study the interpretability of the \textit{D}omain-\textit{A}dapted models via \textit{P}rompt-based \textit{F}ine-tuning (DAPF) framework, which casts dementia detection as diagnosis-related masked-token prediction.  We interpret DAPF and strong baselines using a variety of probing and analysis techniques, finding that  DAPF achieved the best overall performance (accuracy=0.83 and macro-F\textsubscript{1}=0.83) with diagnosis most recoverable from its \texttt{[MASK]} representation.  However, this representational advantage did not extend to token-level explanation faithfulness.  DAPF attributions primarily reflected language task vocabulary, discourse markers, and transcription artifacts, with perturbation tests showing weak or negative effects.  This suggests that its masked-token interface determines diagnosis information without producing faithful token-level explanations. 
\end{abstract}

\section{Introduction}

Detecting Alzheimer's disease and related dementias (ADRD) from spoken language offers a low-cost, non-invasive alternative to traditional biomarkers \cite{de_la_fuente_garcia_artificial_2020, ding_speech_2024, lima_evaluating_2025}. Rather than requiring neuroimaging, cerebrospinal fluid collection, plasma biomarkers, or blood-based assays \cite{khan2015alzheimer, https://doi.org/10.1002/alz.14528, mielke2024recommendations}, spoken dialogue can be elicited through picture description, story recall, or conversational interviews \cite{luz_alzheimers_2020, lanzi_dementiabank_2023, clarke_comparison_2021, pope_finding_2011}.  This makes it a useful medium for studying differences in lexical choice, fluency, discourse organization, or other ADRD-associated language changes \cite{ahmed2013connected, slegers2018connected, gao2024dependency}. 

However, cross-corpus generalization is difficult for spoken language ADRD detection.  Datasets differ in size, participant demographics, and elicitation tasks, and high accuracy on a specific corpus is not conclusive evidence that a model has learned linguistic or task-specific patterns relevant to ADRD \cite{farzana-parde-2023-towards, wang-etal-2022-identifying}.
\textit{D}omain-\textit{A}dapted models via \textit{P}rompt-based \textit{F}ine-tuning (DAPF) frames cross-domain detection as a cloze-style masked-token task by embedding transcripts in prompts containing domain information and diagnosis-related label words \cite{farzana-parde-2024-domain}, predicting label words at a \texttt{[MASK]} position and mapping them to actual labels using a verbalizer. From an interpretability perspective, this poses an interesting case.  Manual prompts are readable, but not necessarily faithful explanations of a model’s decision \cite{jacovi-goldberg-2020-towards, lyu-etal-2024-towards}. Prompt-based classifiers may also rely on the prompt structure and other dataset-specific cues \cite{jiang-etal-2020-know, hu-etal-2022-knowledgeable, farzana-parde-2024-domain}.

In this paper, we perform systematic interpretability experiments on DAPF in a controlled cross-domain setting.  We use representation probing \cite{belinkov-2022-probing} to test whether diagnosis information is recoverable from hidden states at specific positions, and Integrated Gradients \cite{pmlr-v70-sundararajan17a} to assign token-level contributions to the predicted \textsc{Dementia} probability. We aggregate attributions across random seeds by normalized token type, then use deletion and insertion tests \cite{Fong_2017_ICCV, DBLP:conf/bmvc/PetsiukDS18} to evaluate attribution faithfulness and paired error analysis to compare model decisions.

Our findings confirm that many high-attribution tokens reflect the elicitation task or transcription conventions rather than clear ADRD-related language behavior.  They also reveal a split between representation-level interpretability and token-level faithfulness. DAPF achieves higher accuracy and macro-F1 than control baselines, and its final-layer  \texttt{[MASK]} representation has more information than its own \texttt{[CLS]} representation and the final-layer \texttt{[CLS]} representations of the baselines. However, perturbation tests show that DAPF’s highlighted tokens do not consistently support the model’s \textsc{Dementia} prediction.  Thus, while DAPF appears to concentrate diagnosis information at the prompt prediction position, this does not necessarily lead to faithful word-level explanations.

\section{Related Work}

\subsection{Spoken-Language Dementia Detection}
\label{subsec:topic_and_data}
Speech-based ADRD detection has been surveyed extensively \cite{petti_systematic_2020, de_la_fuente_garcia_artificial_2020, qi_noninvasive_2023, ding_speech_2024, slegers2018connected}.  The most popular dataset for the task is DementiaBank, and the Pitt Corpus within it, based on picture descriptions \cite{lanzi_dementiabank_2023}. ADReSS standardized the Pitt corpus, while ADReSSo and ADReSS-M extended it to speech-only detection and multilingual evaluation \cite{luz_alzheimers_2020, luz_detecting_2021, luz_overview_2024}. Other previously used datasets include CCC, which contains longer health-oriented conversational interviews \cite{pope_finding_2011}, and SLaCAD, which links spoken-language data with clinical and biomarker measures for early ADRD detection \cite{farzana-etal-2024-slacad}. Their differences in elicitation task, transcript length, participants, and clinical annotation make them useful for studying cross-domain generalization versus spurious fitting to task-specific vocabulary or dataset artifacts.

\subsection{Prompt Learning and Domain Adaptation}
Prompt learning has been widely used to adapt pretrained language models through manually designed prompts, automatically generated prompts, and learned continuous prompts \cite{schick-schutze-2021-exploiting, shin-etal-2020-autoprompt, gao-etal-2021-making, ding-etal-2022-openprompt, li-liang-2021-prefix, lester-etal-2021-power, liu-etal-2022-p}. It has also been applied to domain adaptation by learning domain-specific prompts while keeping most model parameters fixed \cite{ben-david-etal-2022-pada, goswami-etal-2023-switchprompt, hedderich-etal-2021-survey}. In spoken-language ADRD detection, DAPF uses prompt-based masked-token prediction to improve cross-domain performance \cite{farzana-parde-2024-domain}. While previous work focused on its predictive performance, we study how the model represents diagnosis-related information and whether its predictions can be interpreted reliably.

\subsection{Representation Probing}

Representation probing trains auxiliary classifiers to test what information is recoverable from model's hidden states \cite{belinkov-2022-probing}. Layer-wise probing has been used to examine where linguistic information emerges in BERT \cite{tenney-etal-2019-bert}. Since probe performance can reflect the probe’s own learning capacity, prior work recommends simple probes and control tasks \cite{hewitt-liang-2019-designing}. We therefore use logistic regression and verify the results with a random-label control.

\subsection{Post-hoc Attribution Interpretability}

Post-hoc interpretability methods are widely used to identify which parts of an input contribute to a prediction \cite{10.1145/3546577}. In clinical NLP, a model might appear to perform well based on dataset-specific cues, preprocessing artifacts, or clinically irrelevant tokens \cite{wang-etal-2022-identifying, vasquez-venegas_detecting_2024}.
Gradient-based token-level attribution methods estimate the contribution of input tokens from the model's gradients \cite{simonyan2013deep, ancona_gradient-based_2019, smilkov2017smoothgrad, pmlr-v70-sundararajan17a}; for instance, Integrated Gradients traces how the model’s prediction changes as the input moves from a baseline version to the actual example \cite{pmlr-v70-sundararajan17a}. Perturbation-based methods, such as occlusion, LIME, SHAP, and feature ablation, instead measure how the prediction changes when parts of the input are removed, masked, or replaced \cite{fleet_visualizing_2014, 10.1145/2939672.2939778, lundberg_unified_2017, li2016understanding, Fong_2017_ICCV}. 

Attribution scores should not be treated as faithful explanations by default. Highlighted tokens may look plausible while failing to reflect the model's actual decision process \cite{jacovi-goldberg-2020-towards, poerner-etal-2018-evaluating}. If the highlighted tokens are truly important for ADRD prediction, removing them should weaken that prediction, while retaining or restoring them should help preserve or recover it \cite{Fong_2017_ICCV, DBLP:conf/bmvc/PetsiukDS18, deyoung-etal-2020-eraser}.
Transcript-based models may capture tokens associated with impairment, but they may also exploit cues specific to the task or corpus, such as picture-description content, transcription/disfluency tags, interviewer interactions, or other dataset-specific artifacts \cite{cummings2019describing, farzana-etal-2022-say, heitz-etal-2024-influence,farzana-parde-2023-towards, liu2024clever}. 
We use Integrated Gradients, cross-seed attribution aggregation, and deletion/insertion tests to examine whether DAPF's predictions are supported by stable and faithful token-level evidence.

\subsection{Interpretability in Dementia Detection}

Interpretability is important for building trust in spoken-language ADRD detection systems. Earlier work emphasized hand-crafted acoustic and linguistic features, including lexical, syntactic, semantic, discourse, and fluency measures \cite{fraser_linguistic_2015, mueller_connected_2018, davis_examining_2009, rohanian_alzheimer8217s_2021, petti_systematic_2020, qi_noninvasive_2023}. More recent studies have applied post-hoc explanation methods to ADRD models, identifying clinically plausible linguistic cues while also showing that explanations can be influenced by corpus-specific and preprocessing-related artifacts \cite{shankar_systematic_2025, vimbi_interpreting_2024, iqbal2024explainable, oiza-zapata_alzheimers_2025, wang2020explainable, zhu_towards_2022, gallardo2025explainable, farzana-etal-2022-say, heitz-etal-2024-influence, farzana-parde-2023-towards, liu2024clever}. Most prior work has focused on the clinical plausibility or usefulness of explanations. In contrast, we study whether a prompt-based ADRD detector encodes diagnosis information at the prompt prediction position and whether its token-level attributions are stable and faithful.

\section{Task and Model Background}

\subsection{Task}

We study binary ADRD detection from spoken-language transcripts using one treatment model (DAPF) and two baselines (BERT-CLS and BERT-CLS+Prompt) to distinguish the effects of prompts and special tokens. The models are described in \S{}\ref{subsec:dapf} and \S{}\ref{subsec:baselines}. Given the limited availability of spoken-language ADRD data, we evaluate all models in a cross-domain setting using the Carolina Conversations Collection (CCC) as the source domain \cite{pope_finding_2011} and ADReSS as the target domain \cite{luz_alzheimers_2020}.  Details regarding those datasets were provided in \S{}\ref{subsec:topic_and_data}.

We use the \textit{ADReSS train} split as the target training domain for CCC $\rightarrow$ ADReSS adaptation and evaluate on the held-out \textit{ADReSS test} split. Class distributions are shown in Table~\ref{tab:data_stats}. The small size of both datasets makes interpretability analysis challenging, as token-level explanations can be sensitive to individual participants, repeated task vocabulary, and model initialization.


\begin{table}[t]
\centering
\small
\begin{tabular}{lccc}
\toprule
\textbf{Dataset Split} & \textbf{Total} & \textbf{AD} & \textbf{HC} \\
\midrule
Source Train (CCC) & 105 & 57 & 48 \\
Target Train (ADReSS) & 108 & 54 & 54 \\
Target Test (ADReSS) & 48 & 24 & 24 \\
\bottomrule
\end{tabular}
\caption{Dataset splits. \textit{AD}=Alzheimer's disease or related dementia; \textit{HC}=Healthy control.}
\label{tab:data_stats}
\end{table}

\subsection{Prompt-Based Domain Adaptation}
\label{subsec:dapf}
\textit{D}omain-\textit{A}daptive \textit{P}rompt-Based \textit{F}ine-Tuning (DAPF) recasts ADRD detection as a prompt-based masked language modeling task rather than conventional sequence classification \cite{farzana-parde-2024-domain}.  It introduces two manually-designed prompt components: domain-specific prompt text ($D$), which captures corpus-specific characteristics (e.g., CCC or ADReSS); and class-specific prompt text ($C$), which captures diagnosis-specific cues.  DAPF fine-tunes a pretrained language model to predict masked label words (e.g., \textit{AD} or \textit{control} in place of \texttt{[MASK]}) inserted into these prompts, allowing the model to better align its representations with both the target task and differences between the source and target domain data, improving its cross-corpus generalization abilities.
To ensure close replication, we use \texttt{template\_id=7} proposed by \citet{farzana-parde-2024-domain}:

\begin{quote}
    \texttt{Participant's narration on \{domain\}. \{transcript\} Patient has diagnosis [MASK].}
\end{quote} 

Following the original work, we used BERT \cite{devlin-etal-2019-bert} as the backbone model and prediction head. The manual verbalizer maps words predicted in the masked position to class labels:
\begin{quote}
\textsc{Healthy} $\rightarrow$ \{healthy\}

\textsc{Dementia} $\rightarrow$ \{dementia, alzheimer's, disfluent, disordered\}
\end{quote}

This makes DAPF particularly interesting for interpretability, since predictions are made from a single prompt token rather than an aggregated \texttt{[CLS]} sequence representation. If prompt learning works as intended, diagnosis-relevant information should be encoded at the \texttt{[MASK]} position. However, the same structure introduces interpretability risks, which we investigate in this study.

\subsection{Baselines}
\label{subsec:baselines}

To distinguish DAPF-specific explanation behavior from broader limitations of token attribution, we compare DAPF with two BERT sequence classifier baselines with the same data settings.

\paragraph{BERT-CLS Baseline:}
This is a transcript-only baseline that receives the transcript as input and predicts the class through a standard classification head over the \texttt{[CLS]} representation, functioning as a more conventional sequence classifier:

\begin{quote}
\texttt{[CLS]} \texttt{\{transcript\}} \texttt{[SEP]} $\rightarrow$ classifier head $\rightarrow$ \{\textsc{Healthy}, \textsc{Dementia}\}
\end{quote}
The baseline shares the same \textit{BERT-base-uncased} encoder as DAPF, with other settings kept constant. It neither includes the DAPF prompt template nor a diagnosis \texttt{[MASK]} token, so we probe \texttt{[CLS]} only. 

\paragraph{BERT-CLS+Prompt Baseline:}
To separate the effects of prompt text and masked-token prediction, BERT-CLS+Prompt uses the same prompt as DAPF.
However, unlike DAPF, it does not predict label words at the \texttt{[MASK]} position and does not use a verbalizer. Instead, it is trained using a classification head over \texttt{[CLS]}. Therefore, we  treat its [MASK] representation as a prompt-position artifact rather than a meaningful diagnosis signal. This baseline isolates the effect of the prompt alone, while retaining \texttt{[CLS]} classification.

\section{Interpretability Methodology}
We interpret using representation probing (\S{}\ref{subsec:representation_probing}), token- (\S{}\ref{subsec:integrated_gradients}) and word-level (\S{}\ref{subsec:word_level_attribution}) attributions, and perturbation-based faithfulness diagnostics (\S{}\ref{subsec:faithfulness_tests}).  We also report paired error analyses based on output predictions (\S{}\ref{subsec:paired_error_analysis}).

\subsection{Representation Probing}
\label{subsec:representation_probing}

We evaluate whether diagnosis information is encoded in hidden states using representation probing. For DAPF, we save hidden states at the \texttt{[CLS]} (sequence representation) and diagnosis \texttt{[MASK]} position. For BERT-CLS, no diagnosis \texttt{[MASK]} token exists, so we probe \texttt{[CLS]} only. Since BERT-CLS+Prompt contains both \texttt{[CLS]} and the auxiliary \texttt{[MASK]}, we probe both, although only \texttt{[CLS]} is used for prediction.
For each condition, we train a logistic regression model on frozen hidden states using stratified five-fold cross-validation and macro-F1 scoring to measure how well diagnosis can be recovered from a representation. We also perform layer-wise probing to study where diagnosis information is recoverable across model layers. 

\subsection{Token-Level Attribution}
\label{subsec:integrated_gradients}

To estimate each token's contribution to the predicted probability of ADRD, $P($\textsc{Dementia}$)$, as our primary attribution method we use Integrated Gradients \cite{pmlr-v70-sundararajan17a}. For DAPF, $P($\textsc{Dementia}$)$ is derived from masked-token verbalizer scores at \texttt{[MASK]}. For BERT-CLS and BERT-CLS+Prompt, $P($\textsc{Dementia}$)$ is extracted from the \texttt{[CLS]} classification head. We report attribution aggregates after removing the prompt-scaffold tokens. A positive attribution score means that the token contributes toward the \textsc{Dementia} class, while a negative score means that the token contributes away from it. 
To examine whether the qualitative attribution patterns depend on the explanation method, we apply Partition SHAP \cite{lundberg_unified_2017} to DAPF over all 48 test transcripts.
We explain $P(\textsc{Dementia})$ using
\texttt{max\_evals}=500. Since this analysis covers one seed and is
not followed by the deletion and insertion tests, we treat it as an
attribution-method sensitivity check rather than as a second
faithfulness evaluation.
Attribution values are not probabilities and should not be interpreted as clinical effect sizes. They are computed for individual token occurrences rather than types, making them input-specific; each token occurrence may receive a different score depending on its context, position, surrounding words, and the model’s prediction.

\subsection{Aggregated Token-Level Attribution}
\label{subsec:word_level_attribution}

Since individual token occurrences can be misleading, we use
the same token preprocessing for both attribution methods and
aggregate attributions by normalized token type.  We compute both the mean signed attribution, indicating whether a token tends to support or oppose the \textsc{Dementia} prediction, and the mean absolute attribution, indicating the overall magnitude of its contribution (see Appendix~\ref{appendix:aggregation} for the definitions).
To avoid over-interpreting rare tokens, we analyze tokens that appear at least 10 times and in an average of at least five transcripts. As a sensitivity check, we also evaluated a lower threshold of at least five occurrences and presence in at least three transcripts, which produced similar qualitative results. We therefore report results using the stricter threshold.

\subsection{Faithfulness Tests}
\label{subsec:faithfulness_tests}

\begin{table*}[t]
\centering
\small
\begin{tabular}{p{0.18\textwidth}p{0.23\textwidth}p{0.23\textwidth}p{0.23\textwidth}}
\toprule
\textbf{Component} & \textbf{DAPF-BERT} & \textbf{BERT-CLS} & \textbf{BERT-CLS+Prompt} \\
\midrule
Input format & Prompt-wrapped transcript & Transcript only & Prompt-wrapped transcript \\
Prediction interface & MLM/verbalizer at \texttt{[MASK]} & Classifier over \texttt{[CLS]} & Classifier over \texttt{[CLS]} \\
Template & Manual template ID 7 & - & Manual template ID 7 \\
Verbalizer & Manual & - & - \\
\bottomrule
\end{tabular}
\caption{Model-specific differences between DAPF-BERT, BERT-CLS, and BERT-CLS+Prompt. All models share the same backbone, data split, training configuration, class weighting, and random seeds (Appendix Table~\ref{tab:shared_setup}).}
\label{tab:setup}
\end{table*}

Faithfulness measures whether an explanation reflects the model's actual decision-making process rather than simply providing a plausible rationale \cite{jacovi-goldberg-2020-towards}. We evaluate faithfulness using deletion and insertion tests. In deletion, we mask the highest-ranked tokens and measure the change in the predicted probability of \textsc{Dementia}. In insertion, we begin with those tokens masked, restore them, and measure whether the prediction recovers (we provide formal definitions of these tests in Appendix~\ref{appendix:faithfulness}).

We use two token rankings. The first ranks tokens positively attributed toward \textsc{Dementia}. The second ranks tokens by absolute attribution, regardless of whether they push the prediction toward or away from \textsc{Dementia}. Since Integrated Gradients produces explanations for each transcript, we evaluate faithfulness at the token-occurrence level.

\section{Experimental Setup}

\subsection{Data, Models, and Training}

Table~\ref{tab:setup} summarizes the controlled comparison. All models use the same backbone, data split, training configuration, class weighting, and random seeds. BERT-CLS compares DAPF with a standard \texttt{[CLS]}-based prediction mechanism. The data and backbone were fixed while the \texttt{[MASK]}-based objective was replaced with a standard classification head.
Models are evaluated using balanced accuracy, macro F1, AUROC, and Expected Calibration Error (ECE) \cite{guo2017calibration}; lower ECE indicates better calibration. Full training and interpretability configurations are reported in Appendix Tables~\ref{tab:interp-config} and ~\ref{tab:shared_setup}.

\section{Results}

\subsection{Predictive Performance and Calibration}

Table~\ref{tab:performance-comparison} reports predictive performance (Appendix Table~\ref{tab:performance-comparison-sd}). DAPF achieves the highest accuracy and macro-F1 among the baselines, while BERT-CLS achieves the highest AUROC. ECE is similar across all models.  Since BERT-CLS+Prompt closely matches BERT-CLS in accuracy and macro-F1, adding the DAPF-style prompt does not reproduce DAPF's behavior.
The remaining analyses examine whether DAPF's masked-token prediction objective changes its diagnosis encoding location and its token attribution faithfulness.

\begin{table}[t]
\centering
\small
\renewcommand{\arraystretch}{1.1}
\begin{tabular}{L{1.5cm}C{1cm}C{1cm}C{1cm}C{1cm}}
\toprule
\textbf{Model} & \textbf{Acc.} & \textbf{Macro-F1} & \textbf{AUROC} & \textbf{ECE} \\
\midrule
DAPF &
\textbf{0.8292} &
\textbf{0.8287} &
0.8757 &
0.1683 \\

BERT-CLS &
0.8083 &
0.8063 &
\textbf{0.8906} &
0.1695\\

BERT-CLS+Prompt &
0.8083 &
0.8065 &
0.8819 &
\textbf{0.1680} \\
\bottomrule
\end{tabular}
\caption{Mean test performance over five random seeds.}
\label{tab:performance-comparison}
\end{table}

\subsection{Paired Error Analysis}
\label{subsec:paired_error_analysis}

We compare predictions at the seed-sample level, using the 48 test samples across five random seeds ($n$=240 decisions). Table~\ref{tab:paired-error} summarizes paired-error patterns. All models agree on 170 correct and 19 incorrect cases. DAPF is the only correct model in 20 cases, compared with three for BERT-CLS and two for BERT-CLS+Prompt. However, the classifier baselines are both correct in 17 cases where DAPF is wrong.  This suggests a modest but distinct decision-making advantage for DAPF.

\begin{table}[t]
\centering
\small
\begin{tabular}{lr}
\toprule
\textbf{Correct Model Combination} & \textbf{Count} \\
\midrule
All three correct & 170 \\
All three incorrect & 19 \\
\midrule
DAPF only & 20 \\
BERT-CLS only & 3 \\
BERT-CLS+Prompt only & 2 \\
\midrule
BERT-CLS + BERT-CLS+Prompt & 17 \\
DAPF + BERT-CLS+Prompt & 5 \\
DAPF + BERT-CLS & 4 \\
\bottomrule
\end{tabular}
\caption{Paired error analysis over 48 ADReSS test samples $\times$ five random seeds ($n$=240 decisions). Counts show which model combinations were correct.}
\label{tab:paired-error}
\end{table}

\subsection{Representation Probing}

Table~\ref{tab:final-probes} reports final-layer probing (Appendix Table~\ref{tab:final-probes-sd}). DAPF's  \texttt{[MASK]} representation is the most diagnostic final-layer representation, outperforming DAPF \texttt{[CLS]} across seeds, with a mean paired macro-F1 improvement of 0.044 (bootstrap 95\% CI: [0.034, 0.056]). A one-sided exact Wilcoxon signed-rank test confirmed that the supervised \texttt{[MASK]} representation contains more diagnosis information than \texttt{[CLS]} (p=0.031). BERT-CLS \texttt{[CLS]} is slightly stronger than DAPF \texttt{[CLS]}, but weaker than DAPF \texttt{[MASK]}. BERT-CLS+Prompt's final-layer \texttt{[CLS]} probe score is close to BERT-CLS, while its auxiliary \texttt{[MASK]} representation also contains diagnosis information, despite not being used for prediction. This suggests that the prompt \texttt{[MASK]} token accumulates diagnosis-relevant contextual information, but its auxiliary \texttt{[MASK]} representation is less informative than DAPF's supervised diagnosis \texttt{[MASK]}, indicating that DAPF's masked-token objective further specializes the prompt prediction position.

\begin{table}[t]
\centering
\small
\renewcommand{\arraystretch}{1.1}
\begin{tabular}{llC{1.35cm}}
\toprule
\textbf{Model} & \textbf{Representation} & \textbf{Probe Macro-F1} \\
\midrule
DAPF & \texttt{[CLS]} & 0.7550  \\
\textbf{DAPF} & \textbf{\texttt{[MASK]}} & \textbf{0.7989} \\
BERT-CLS & \texttt{[CLS]} & 0.7704 \\
BERT-CLS+Prompt & \texttt{[CLS]} & 0.7683\\
BERT-CLS+Prompt & auxiliary \texttt{[MASK]} & 0.7662\\
\bottomrule
\end{tabular}
\caption{Final-layer representation probing. BERT-CLS has no diagnosis \texttt{[MASK]}. For BERT-CLS+Prompt, \texttt{[MASK]} is auxiliary and not used for prediction.}
\label{tab:final-probes}
\end{table}

\begin{table}[t]
\centering
\small
\begin{tabular}{lccc}
\toprule
\textbf{Representation} & \textbf{Diagnosis Probe} & \textbf{Random Probe}  \\
\midrule
DAPF [CLS]  & 0.7550 $\pm$ 0.0113 & 0.474 $\pm$ 0.007  \\
DAPF [MASK] & 0.7989 $\pm$ 0.0189 & 0.470 $\pm$ 0.004 \\
\bottomrule
\end{tabular}
\caption{Random-label control, with macro-F1 averaged across five random seeds. Random probes use the same DAPF representations and probing procedure but randomly permute diagnosis labels.}
\label{tab:random-label-probe}
\end{table}

As a sanity check, we repeated the DAPF probing after randomly shuffling diagnosis labels, while keeping the hidden representations and class balance unchanged. Probe macro-F1 performance dropped from 0.755 to 0.474 for \texttt{[CLS]} and from 0.799 to 0.470 for \texttt{[MASK]} (Table~\ref{tab:random-label-probe}), indicating that the probes recover diagnosis-related rather than arbitrary structure.

\subsection{Layer-wise Representation Probing}

\begin{figure}
    \centering
    \begin{tikzpicture}
\begin{axis}[
    width=\columnwidth,
    height=5.5cm,
    xlabel={Layer},
    ylabel={Macro-F1},
    xmin=0.5, xmax=12.5,
    ymin=0.48, ymax=0.87,
    xtick={1,2,...,12},
    ytick={0.50,0.55,0.60,0.65,0.70,0.75,0.80,0.85},
    yticklabel={\pgfmathprintnumber[fixed, fixed zerofill, precision=2]{\tick}},
    yticklabel style={font=\footnotesize},
    xticklabel style={font=\footnotesize},
    xlabel style={
        font=\small,
        at={(axis description cs:0.5,-0.10)},
        anchor=north
    },
    ylabel style={font=\small},
    legend style={
        at={(rel axis cs:0.98,0.03)},
        anchor=south east,
        legend columns=1,
        font=\scriptsize,
        draw=gray!60,
        fill=white,
        fill opacity=0.90,
        text opacity=1,
        inner xsep=2pt,
        inner ysep=1pt,
        row sep=-2pt,
        column sep=2pt,
        /tikz/every even column/.append style={column sep=2pt},
    },
    legend image post style={xscale=0.6},
    grid=both,
    minor grid style={line width=0.2pt, draw=gray!15},
    major grid style={line width=0.5pt, draw=gray!35},
    minor tick num=1,
]
 
\addplot[color=blue!80!black, mark=circle*, mark size=1.2pt, line width=1.4pt]
coordinates {
    (1,0.7101)+-(0,0.0102) (2,0.6976)+-(0,0.0216) (3,0.7644)+-(0,0.0116)
    (4,0.6746)+-(0,0.0132) (5,0.7488)+-(0,0.0204) (6,0.7490)+-(0,0.0375)
    (7,0.7201)+-(0,0.0264) (8,0.7472)+-(0,0.0260) (9,0.7945)+-(0,0.0200)
    (10,0.7733)+-(0,0.0457) (11,0.7567)+-(0,0.0187) (12,0.7550)+-(0,0.0113)
};
\addlegendentry{DAPF \texttt{[CLS]}}
 
\addplot[color=red!75!black, mark=square*, mark size=1.2pt, line width=1.4pt]
coordinates {
    (1,0.5654)+-(0,0.0107) (2,0.6009)+-(0,0.0183) (3,0.6306)+-(0,0.0266)
    (4,0.7439)+-(0,0.0106) (5,0.7916)+-(0,0.0315) (6,0.7361)+-(0,0.0175)
    (7,0.7176)+-(0,0.0514) (8,0.6975)+-(0,0.0245) (9,0.7119)+-(0,0.0227)
    (10,0.7497)+-(0,0.0403) (11,0.7760)+-(0,0.0343) (12,0.7989)+-(0,0.0189)
};
\addlegendentry{DAPF \texttt{[MASK]}}
 
\addplot[color=green!55!black, mark=triangle*, mark size=1.2pt, line width=1.4pt]
coordinates {
    (1,0.7138)+-(0,0.0148) (2,0.7860)+-(0,0.0130) (3,0.7419)+-(0,0.0142)
    (4,0.7979)+-(0,0.0150) (5,0.8040)+-(0,0.0214) (6,0.7900)+-(0,0.0193)
    (7,0.7463)+-(0,0.0325) (8,0.7585)+-(0,0.0178) (9,0.7803)+-(0,0.0545)
    (10,0.7653)+-(0,0.0354) (11,0.7543)+-(0,0.0363) (12,0.7704)+-(0,0.0260)
};
\addlegendentry{BERT-CLS \texttt{[CLS]}}
 
\addplot[color=orange!85!black, mark=diamond*, mark size=1.2pt, line width=1.4pt]
coordinates {
    (1,0.7802)+-(0,0.0283) (2,0.6911)+-(0,0.0091) (3,0.7379)+-(0,0.0405)
    (4,0.7277)+-(0,0.0274) (5,0.7498)+-(0,0.0304) (6,0.7924)+-(0,0.0289)
    (7,0.7558)+-(0,0.0308) (8,0.7760)+-(0,0.0285) (9,0.7759)+-(0,0.0214)
    (10,0.7436)+-(0,0.0484) (11,0.7496)+-(0,0.0163) (12,0.7683)+-(0,0.0152)
};
\addlegendentry{BERT-CLS+Prompt \texttt{[CLS]}}
 
\addplot[color=violet!85!black, mark=pentagon*, mark size=1.2pt, line width=1.4pt]
coordinates {
    (1,0.5437)+-(0,0.0093) (2,0.5396)+-(0,0.0143) (3,0.6439)+-(0,0.0380)
    (4,0.6773)+-(0,0.0233) (5,0.6922)+-(0,0.0481) (6,0.7466)+-(0,0.0511)
    (7,0.7261)+-(0,0.0395) (8,0.7837)+-(0,0.0342) (9,0.7605)+-(0,0.0322)
    (10,0.7833)+-(0,0.0530) (11,0.7752)+-(0,0.0469) (12,0.7662)+-(0,0.0374)
};
\addlegendentry{BERT-CLS+Prompt aux.~\texttt{[MASK]}}
 
\end{axis}
\end{tikzpicture}
    \caption{Layer-wise probing across models.  Macro-F1 is averaged across five random seeds.}
    \label{fig:layer-wise_probing}
\end{figure}
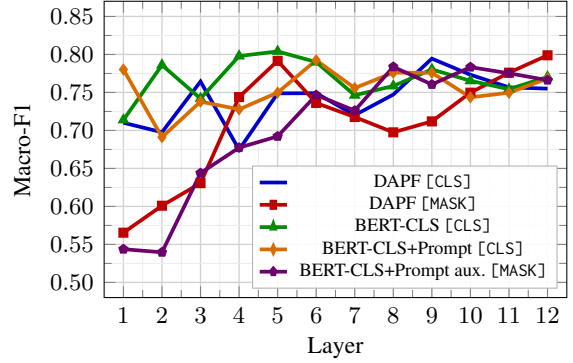

Figure~\ref{fig:layer-wise_probing} shows layer-wise probing results (Appendix Table~\ref{tab:layerwise-probing}). DAPF's \texttt{[MASK]} representation becomes informative in the middle layers and peaks in the final layer. BERT-CLS \texttt{[CLS]} peaks around layer 5, suggesting that middle-layer representations contain useful diagnosis information, while DAPF concentrates it at the prompt prediction position.\footnote{Layer 4-5 results should be read cautiously, since they may encode lexical or task-related differences between groups; probing does not reveal what drives the separation. Future work could investigate random-label controls and probe nuisance variables (e.g., transcript length or number of task-related content words).} The auxiliary \texttt{[MASK]} in BERT-CLS+Prompt is weak in early layers but becomes more diagnosis-informative after incorporating transcript context in subsequent layers, reaching its strongest values around layers 8–10 rather than the final layer. Thus, it appears that the prompt-as-input setting can produce a diagnostically informative auxiliary prompt position while supervised masked-token prediction further specializes the final \texttt{[MASK]} representation.

\subsection{DAPF Attribution Patterns and Method Sensitivity}

Across five seeds, 163 normalized token types met the frequency threshold. Signed attributions were centered near zero, mean absolute attribution was modest, and positive and negative tokens were nearly balanced (see Appendix Table~\ref{tab:attr-summary-percentile} for specific values).
Table~\ref{tab:top-positive} shows the highest positive-attribution tokens. These include Cookie Theft vocabulary (e.g., \texttt{reach}, \texttt{cookies}) together with generic discourse words (e.g., \texttt{something}, \texttt{well}), suggesting that DAPF emphasizes task-specific vocabulary and conversational markers rather than clearly interpretable ADRD-related features.

\begin{table}[t]
\centering
\small
\begin{tabular}{lrrccc}
\toprule
\textbf{Token} & \textbf{Total} & \textbf{T/S} & \textbf{Attr.} & \textbf{SD} & \textbf{|Attr.|} \\
\midrule
reach & 55 & 8 & 0.2540 & 0.1900 & 0.3374 \\
out & 240 & 32 & 0.1832 & 0.1569 & 0.2337 \\
something & 25 & 5 & 0.1827 & 0.0946 & 0.2568 \\
anything & 40 & 6 & 0.1262 & 0.0808 & 0.1584 \\
cookies & 150 & 21 & 0.1130 & 0.1906 & 0.2982 \\
quiet & 25 & 5 & 0.0921 & 0.1086 & 0.1211 \\
well & 145 & 21 & 0.0717 & 0.0298 & 0.1149 \\
nice & 35 & 5 & 0.0657 & 0.0312 & 0.0915 \\
no & 45 & 7 & 0.0630 & 0.0486 & 0.0860 \\
by & 25 & 5 & 0.0585 & 0.0553 & 0.0799 \\
\bottomrule
\end{tabular}
\caption{Top positive DAPF cross-seed mean attribution tokens. Positive values indicate average contribution toward the \textsc{Dementia} class. \textit{Total}=total occurrences; \textit{T/S}=mean transcripts per seed;  \textit{Attr.}=mean signed attribution; \textit{SD}=standard deviation across seeds;  \textit{|Attr.|}=mean absolute attribution.}
\label{tab:top-positive}
\end{table}

Table~\ref{tab:top-negative} shows a similar pattern, including task vocabulary, discourse words, fillers, and transcription artifacts. Although fillers can be associated with cognitive impairment \cite{farzana-etal-2022-say}, attribution alone does not justify interpreting \texttt{uh} as evidence against ADRD. Overall, the aggregated attributions are more stable across seeds but remain clinically difficult to interpret.

\begin{table}[t]
\centering
\small
\begin{tabular}{lrrccc}
\toprule
\textbf{Token} & \textbf{Total} & \textbf{T/S} & \textbf{Attr.} & \textbf{SD} & \textbf{|Attr.|} \\
\midrule
okay & 60 & 12 & -0.5181 & 0.3719 & 0.5775 \\
lid & 40 & 7 & -0.1401 & 0.0474 & 0.1456 \\
guess & 45 & 8 & -0.1141 & 0.0574 & 0.1407 \\
cabinets & 35 & 5 & -0.1017 & 0.0865 & 0.1366 \\
curtains & 55 & 11 & -0.0977 & 0.0196 & 0.1726 \\
laughing & 35 & 7 & -0.0813 & 0.0523 & 0.1502 \\
washing & 55 & 9 & -0.0780 & 0.0283 & 0.1133 \\
uh & 635 & 41 & -0.0775 & 0.0240 & 0.1171 \\
off & 145 & 20 & -0.0665 & 0.0303 & 0.0997 \\
xx & 95 & 11 & -0.0665 & 0.0547 & 0.1555 \\
\bottomrule
\end{tabular}
\caption{Top negative DAPF cross-seed mean attribution tokens. Negative values indicate average contribution away from the \textsc{Dementia} class.  \textit{Total}=total occurrences; \textit{T/S}=mean transcripts per seed; \textit{Attr.}=mean signed attribution; \textit{SD}=standard deviation across seeds; \textit{|Attr.|}=mean absolute attribution.}
\label{tab:top-negative}
\end{table}

Partition SHAP produced a similar overall pattern. High-attribution items primarily included Cookie Theft vocabulary, such as \textit{fall}, \textit{reach}, \textit{cookie}, and \textit{dish}; discourse forms, such as \textit{well}, \textit{oh}, and \textit{but}; and longer token fragments, such as \textit{ing} and \textit{tin}. Agreement on individual tokens was only partial. Among the 20 tokens reported in the Integrated Gradients top-positive and top-negative tables, 11 (55\%) had the same mean direction under both methods, with a descriptive Spearman correlation of $\rho=0.362$. Overall, the two methods highlight similar types of cues, but do not consistently agree on which words are most important.

\subsection{Faithfulness Comparison}
\label{subsec:faithfulness_comparison}

Table~\ref{tab:faith-positive-compare} compares perturbation-based faithfulness for tokens ranked by positive attribution toward \textsc{Dementia} (Appendix Table~\ref{tab:faith-positive-compare-sd}).
 DAPF shows weak deletion effects and consistently negative insertion effects. In contrast, both BERT-CLS and BERT-CLS+Prompt exhibit positive deletion and steadily increasing insertion scores.

\begin{table}[t]
\centering
\small
\renewcommand{\arraystretch}{1.1}
\begin{tabular}{L{1.45cm}L{1.1cm}C{1cm}C{1cm}C{1cm}}
\toprule
\textbf{Model} & \textbf{Test} & \textbf{5\%} & \textbf{20\%} & \textbf{50\%} \\
\midrule
DAPF & Deletion & 0.0080 & 0.0055  & -0.0318  \\
DAPF & Insertion  & -0.0187 & -0.0352  & -0.0916 \\
\midrule
BERT-CLS & Deletion & 0.0259 & 0.0142 & 0.0067  \\
BERT-CLS & Insertion  & 0.0369 & 0.0813  & 0.1223 \\
\midrule
BERT-CLS +Prompt & Deletion & 0.0302 & 0.0426 & 0.0259 \\
BERT-CLS +Prompt & Insertion & 0.0279  & 0.0787 & 0.1225 \\
\bottomrule
\end{tabular}
\caption{Positive-ranking faithfulness comparison using tokens ranked by positive attribution toward \textsc{Dementia}. Positive values indicate behavior consistent with dementia-supporting evidence. Results are shown for 5\%, 20\%, and 50\% perturbation fractions.}
\label{tab:faith-positive-compare}
\end{table}

Although DAPF has the strongest final-layer diagnosis representation, its positively attributed tokens do not recover the predicted probability of \textsc{Dementia} under insertion. In contrast, BERT-CLS shows more responsive positive attributions under insertion. Likewise, BERT-CLS+Prompt shows positive insertion effects, suggesting that prompt text alone is not responsible for DAPF’s weak insertion faithfulness. The masked-token/verbalizer prediction setup or the attribution behavior when the decision depends on the prompt prediction position are likelier culprits.

We also evaluated faithfulness using tokens ranked by absolute attribution and found that the overall pattern was similar (Appendix~\ref{app:faithfulness}).
Overall, DAPF combines a strongly diagnostic supervised
\texttt{[MASK]} representation with weak positive-attribution
faithfulness. Since BERT-CLS+Prompt does not show the same
insertion pattern, the difference is unlikely to result from
prompt text alone.

\section{Discussion}

\paragraph{Representation-level encoding and token-level faithfulness diverge.}
Classification and probing results show that DAPF learned meaningful diagnosis-related information, with predictive performance slightly higher than BERT-CLS. The final-layer \texttt{[MASK]} representation was consistently more diagnosis-informative than DAPF's \texttt{[CLS]} representation, supporting the hypothesis that diagnosis information is concentrated at the supervised prompt prediction position.
However, faithfulness tests show that BERT-CLS positive attributions are more responsive under insertion than DAPF's, uncovering a mismatch between representation-level diagnosis encoding and token-level faithfulness.

\paragraph{Why are DAPF attributions less convincing?}
Predicting ADRD through the prompt \texttt{[MASK]} position combines information from the transcript, prompt wording, domain phrase, and verbalizer words, creating challenges in explaining why positive attributions fail under insertion. One possibility is that these interactions are not well localized to
individual words. Partition SHAP reproduced the same broad cue categories but agreed only partially with Integrated Gradients on the direction and ranking of individual tokens. This does not by itself
establish unfaithfulness, but it suggests that the word-level explanation is sensitive to the attribution method. BERT-CLS has a simpler prediction path, mapping \texttt{[CLS]} representations directly to class logits, which may better align its positive attributions with the insertion test.
BERT-CLS+Prompt uses the same prompt-wrapped input as DAPF but predicts through a \texttt{[CLS]} head, resulting in behavior closer to BERT-CLS in aggregate performance and paired-error patterns and indicating that DAPF's differences are not explained by prompt text alone.  Further controls could isolate the roles of the masked-language-model head, the manual verbalizer, and the task/domain prompt.

\paragraph{Prompt-as-input control.}
BERT-CLS+Prompt behaved more like BERT-CLS than DAPF in
predictive performance and attribution faithfulness, while
its auxiliary \texttt{[MASK]} representation was less
diagnostic than DAPF's supervised \texttt{[MASK]}. Therefore,
prompt wording alone does not explain DAPF's representation
or faithfulness patterns.

\paragraph{Implications for ADRD detection.}
Future interpretability work should look beyond token attributions to broader transcript patterns, such as informativeness, lexical specificity, and discourse organization. It could also test targeted changes to transcripts based on clinically meaningful features rather than high-attribution tokens. Concept-based methods could help determine whether these patterns are encoded in the model’s hidden states. More broadly, these findings suggest that interpretability should be considered alongside performance when evaluating low-resource prompt-based models, since strong performance alone does not guarantee faithful explanations.

\section{Conclusion}

In this paper, we investigated the interpretability of a BERT-based DAPF model for ADRD detection against transcript-only (BERT-CLS) and prompt-as-input (BERT-CLS+Prompt) baselines. While DAPF achieves slightly higher accuracy and unique correct decisions by concentrating diagnosis information at its supervised \texttt{[MASK]} representation, our analyses show that representation-level encoding and token-level faithfulness diverge. Cross-seed Integrated Gradients showed that high-attribution tokens primarily reflect task-specific vocabulary, discourse markers, and transcription artifacts rather than clinically meaningful ADRD markers. Partition SHAP sensitivity check recovered the same broad cue categories, but showed only partial agreement with Integrated Gradients on the direction and ranking of individual tokens. Furthermore, perturbation tests showed that DAPF's token attributions are less faithful under deletion and insertion than the classifier baselines, indicating that masked-token prompting does not automatically guarantee attribution validity. Ultimately, these findings highlight a mismatch between diagnosis localization and explanation reliability, cautioning that prompt-based model explanations should not be interpreted as faithful clinical rationales.

\section*{Limitations}
All experiments use a single low-resource CCC $\rightarrow$ ADReSS transfer setting, so the findings should not be taken as general claims about speech-based ADRD detection. Since masking large portions of a transcript may create out-of-distribution inputs, effects at larger perturbation fractions, particularly 50\%, may reflect both token importance and sensitivity to distribution shift \cite{hase_out--distribution_2021}.  In the BERT-CLS+Prompt control, we test one DAPF-style prompt; it is possible that different prompt templates, domain descriptions, verbalizers, or prompt-tuning methods may produce different representation and attribution patterns.
Finally, representation-level interpretability does not necessarily mean faithful word-level explanations. In DAPF, the \texttt{[MASK]} representation contains strong diagnosis information, but the token attributions do not behave like word-level evidence under perturbation. This limits how strongly the highlighted tokens can be interpreted as explanations for the model's decisions.

\section*{Ethical Considerations}
Our experiments use the Carolina Conversations Collection (CCC) and ADReSS datasets, both of which were obtained through their established data access procedures. Access to CCC required institutional ethics review at both our institution and the institution hosting the dataset,\footnote{\url{carolinaconversations.musc.edu/ccc/about/}} which we obtained. Access to ADReSS required approval through the DementiaBank consortium;\footnote{\url{talkbank.org/dementia/ADReSS-2020/index.html}} following that, we additionally obtained an ethics determination at our own institution before using the data. All experiments and reporting of results comply with the datasets' access agreements and terms of use.

It's also worth mentioning that these models are intended for research rather than clinical diagnosis. Token attributions that reflect task-specific vocabulary or transcription artifacts may be misleading if interpreted as clinical evidence, while differences across datasets and participant populations may limit generalization.

\bibliography{custom}
\clearpage

\appendix

\section{Aggregation of Token-Level Attribution}
\label{appendix:aggregation}

For each normalized token type $w$, let $a_i(w)$ denote the attribution assigned to its $i$-th occurrence. We compute the mean signed attribution

\[
\bar{a}(w) = \frac{1}{N_w}\sum_{i=1}^{N_w} a_i(w)
\]

and the mean absolute attribution

\[
\overline{|a|}(w) = \frac{1}{N_w}\sum_{i=1}^{N_w}|a_i(w)|.
\]

The mean signed attribution indicates whether a token tends to support or oppose the \textsc{Dementia} prediction, whereas the mean absolute attribution measures the average magnitude of its contribution regardless of direction.

\section{Faithfulness Metrics}
\label{appendix:faithfulness}

For deletion, we mask the selected high-attribution tokens and compute

\[
\Delta_{\mathrm{del}}
=
P_{\mathrm{original}}
-
P_{\mathrm{mask}},
\]

where \(P_{\mathrm{original}}\) is the original predicted probability of \textsc{Dementia} and \(P_{\mathrm{mask}}\) is the probability after masking the selected tokens. If positively attributed tokens provide faithful evidence for the prediction, \(\Delta_{\mathrm{del}}\) should be positive.

For insertion, we begin from an input in which the selected tokens are masked and compute

\[
\Delta_{\mathrm{ins}}
=
P_{\mathrm{restored}}
-
P_{\mathrm{masked\ start}},
\]

where \(P_{\mathrm{masked\ start}}\) is the predicted probability for the masked input and \(P_{\mathrm{restored}}\) is the probability after restoring the selected tokens. Faithful positively attributed tokens should produce a positive \(\Delta_{\mathrm{ins}}\).

\subsection{Interpretability Configuration}

Table~\ref{tab:interp-config} summarizes the interpretability settings. Representation probing tests whether diagnosis can be recovered from frozen hidden states, while Integrated Gradients assigns token-level contributions to $P($\textsc{Dementia}$)$. Since attribution scores are assigned to individual token occurrences, we aggregate normalized token types to identify patterns that are more stable across seeds. Faithfulness is then evaluated with deletion and insertion tests using positive and absolute attribution rankings. We evaluate multiple perturbation fractions to compare local effects with behavior under stronger perturbation; the 50\% condition is included as a stress test rather than a precise estimate of token importance.

\begin{table}[t]
\centering
\small
\begin{tabular}{ll}
\toprule
\textbf{Analysis} & \textbf{Configuration} \\
\midrule
Representation probing & Logistic regression probe \\
Probe evaluation & Stratified 5-fold CV \\
Probe metric & Macro-F1 \\
Primary attribution method & Integrated Gradients \\
Attribution sensitivity check & Partition SHAP (DAPF seed 0) \\
Attribution target & $P(\mathrm{\textsc{Dementia}})$ \\
Attribution unit & Token occurrence \\
Aggregation unit & Normalized token type \\
Main aggregation threshold & Count $\geq$ 10, samples $\geq$ 5 \\
Faithfulness rankings & Positive, absolute \\
Perturbation fractions & 5\%, 10\%, 20\%, 30\%, 50\% \\
\bottomrule
\end{tabular}
\caption{Interpretability configuration.}
\label{tab:interp-config}
\end{table}

\section{Mean test performance over five random seeds with Standard Deviations}
\label{subsec:Mean_test_performance}

Table~\ref{tab:performance-comparison-sd} extends Table~\ref{tab:performance-comparison} by reporting the mean performance together with standard deviations across the five random seeds.

\begin{table}[t]
\centering
\small
\renewcommand{\arraystretch}{1.1}
\begin{tabular}{L{1.5cm}C{1cm}C{1cm}C{1cm}C{1cm}}
\toprule
\textbf{Model} & \textbf{Acc.} & \textbf{Macro-F1} & \textbf{AUROC} & \textbf{ECE} \\
\midrule
DAPF &
\textbf{0.8292 \scriptsize{$\pm$ 0.0174}} &
\textbf{0.8287 \scriptsize{$\pm$ 0.0173}} &
0.8757 \scriptsize{$\pm$ 0.0183} &
0.1683 \scriptsize{$\pm$ 0.0284} \\

BERT-CLS &
0.8083 \scriptsize{$\pm$ 0.0243} &
0.8063 \scriptsize{$\pm$ 0.0247} &
\textbf{0.8906 \scriptsize{$\pm$ 0.0102}} &
0.1695 \scriptsize{$\pm$ 0.0349} \\

BERT-CLS+Prompt &
0.8083 \scriptsize{$\pm$ 0.0083} &
0.8065 \scriptsize{$\pm$ 0.0082} &
0.8819 \scriptsize{$\pm$ 0.0156} &
\textbf{0.1680 \scriptsize{$\pm$ 0.0476}} \\
\bottomrule
\end{tabular}
\caption{Mean test performance over five random seeds.}
\label{tab:performance-comparison-sd}
\end{table}

\section{Final-layer Representation Probing Results With Standard Deviation} 
Table~\ref{tab:final-probes-sd} extends Table~\ref{tab:final-probes} by reporting standard deviations across the five random seeds.

\begin{table}[t]
\centering
\small
\renewcommand{\arraystretch}{1.1}
\begin{tabular}{llC{1.35cm}}
\toprule
\textbf{Model} & \textbf{Representation} & \textbf{Probe Macro-F1} \\
\midrule
DAPF & \texttt{[CLS]} & 0.7550 \scriptsize{$\quad\pm$ 0.0113} \\
\textbf{DAPF} & \textbf{\texttt{[MASK]}} & \textbf{0.7989 \scriptsize{$\quad\pm$ 0.0189}} \\
BERT-CLS & \texttt{[CLS]} & 0.7704 \scriptsize{$\quad\pm$ 0.0260} \\
BERT-CLS+Prompt & \texttt{[CLS]} & 0.7683 \scriptsize{$\quad\pm$ 0.0152}\\
BERT-CLS+Prompt & auxiliary \texttt{[MASK]} & 0.7662 \scriptsize{$\quad\pm$ 0.0374} \\
\bottomrule
\end{tabular}
\caption{Final-layer representation probing. BERT-CLS has no diagnosis \texttt{[MASK]}. For BERT-CLS+Prompt, \texttt{[MASK]} is auxiliary and not used for prediction.}
\label{tab:final-probes-sd}
\end{table}

\section{Full Layer-Wise Representation Probing Results}

In Table~\ref{tab:layerwise-probing} we report full layer-wise diagnosis probing results across DAPF, transcript-only BERT-CLS, and prompt-as-input BERT-CLS+Prompt. Values are macro-F1 scores averaged across five random seeds. For DAPF, \texttt{[MASK]} is the supervised masked-token diagnosis prediction position. For BERT-CLS+Prompt, the auxiliary \texttt{[MASK]} token is present in the input but is not the supervised prediction site; prediction is made through a standard \texttt{[CLS]} classification head.

\begin{table*}[t]
\centering
\small
\begin{tabular}{rcccC{2.5cm}C{2.5cm}}
\toprule
Layer 
& DAPF \texttt{[CLS]} 
& DAPF \texttt{[MASK]} 
& BERT-CLS \texttt{[CLS]} 
& BERT-CLS+Prompt \texttt{[CLS]} 
& BERT-CLS+Prompt aux.~\texttt{[MASK]} \\
\midrule
1  & 0.7101 $\pm$ 0.0102 & 0.5654 $\pm$ 0.0107 & 0.7138 $\pm$ 0.0148 & 0.7802 $\pm$ 0.0283 & 0.5437 $\pm$ 0.0093 \\
2  & 0.6976 $\pm$ 0.0216 & 0.6009 $\pm$ 0.0183 & 0.7860 $\pm$ 0.0130 & 0.6911 $\pm$ 0.0091 & 0.5396 $\pm$ 0.0143 \\
3  & 0.7644 $\pm$ 0.0116 & 0.6306 $\pm$ 0.0266 & 0.7419 $\pm$ 0.0142 & 0.7379 $\pm$ 0.0405 & 0.6439 $\pm$ 0.0380 \\
4  & 0.6746 $\pm$ 0.0132 & 0.7439 $\pm$ 0.0106 & 0.7979 $\pm$ 0.0150 & 0.7277 $\pm$ 0.0274 & 0.6773 $\pm$ 0.0233 \\
5  & 0.7488 $\pm$ 0.0204 & \textit{0.7916 $\pm$ 0.0315} & \textbf{0.8040 $\pm$ 0.0214} & 0.7498 $\pm$ 0.0304 & 0.6922 $\pm$ 0.0481 \\
6  & 0.7490 $\pm$ 0.0375 & 0.7361 $\pm$ 0.0175 & 0.7900 $\pm$ 0.0193 & \textbf{0.7924 $\pm$ 0.0289} & 0.7466 $\pm$ 0.0511 \\
7  & 0.7201 $\pm$ 0.0264 & 0.7176 $\pm$ 0.0514 & 0.7463 $\pm$ 0.0325 & 0.7558 $\pm$ 0.0308 & 0.7261 $\pm$ 0.0395 \\
8  & 0.7472 $\pm$ 0.0260 & 0.6975 $\pm$ 0.0245 & 0.7585 $\pm$ 0.0178 & 0.7760 $\pm$ 0.0285 & 0.7837 $\pm$ 0.0342 \\
9  & \textbf{0.7945 $\pm$ 0.0200} & 0.7119 $\pm$ 0.0227 & 0.7803 $\pm$ 0.0545 & 0.7759 $\pm$ 0.0214 & 0.7605 $\pm$ 0.0322 \\
10 & 0.7733 $\pm$ 0.0457 & 0.7497 $\pm$ 0.0403 & 0.7653 $\pm$ 0.0354 & 0.7436 $\pm$ 0.0484 & \textbf{0.7833 $\pm$ 0.0530} \\
11 & 0.7567 $\pm$ 0.0187 & 0.7760 $\pm$ 0.0343 & 0.7543 $\pm$ 0.0363 & 0.7496 $\pm$ 0.0163 & 0.7752 $\pm$ 0.0469 \\
12 & 0.7550 $\pm$ 0.0113 & \textbf{0.7989 $\pm$ 0.0189} & 0.7704 $\pm$ 0.0260 & 0.7683 $\pm$ 0.0152 & 0.7662 $\pm$ 0.0374 \\
\bottomrule
\end{tabular}
\caption{Layer-wise diagnosis probing across DAPF, transcript-only BERT-CLS, and prompt-as-input BERT-CLS+Prompt. Values are macro-F1 scores averaged across five random seeds. For DAPF, \texttt{[MASK]} is the supervised masked-token diagnosis prediction position. For BERT-CLS+Prompt, the auxiliary \texttt{[MASK]} token is present in the input but is not the supervised prediction site; prediction is made through a standard \texttt{[CLS]} classification head.}
\label{tab:layerwise-probing}
\end{table*}

\section{Summary statistics for aggregated DAPF token attributions}
\label{app:attribution-statistics}

Table~\ref{tab:attr-summary-percentile} shows  statistics for aggregated DAPF token attributions.

\begin{table}[t]
\centering
\small
\begin{tabular}{lC{2cm}C{2cm}}
\toprule
\textbf{Statistic} & \textbf{Mean Signed Attribution} & \textbf{Mean Abs. Attribution} \\
\midrule
Mean & -0.0015 & 0.0724 \\
Standard Dev. & 0.0608 & 0.0637 \\ \midrule
Minimum & -0.5181 & 0.0177 \\
25th Percentile & -0.0174 & 0.0427 \\
Median & 0.0000 & 0.0540 \\
75th Percentile & 0.0147 & 0.0753 \\
Maximum & 0.2540 & 0.5775 \\
\bottomrule
\end{tabular}
\caption{Summary statistics for aggregated DAPF token attributions ($n=163$).}
\label{tab:attr-summary-percentile}
\end{table}

\section{Experimental Configuration}
\label{app:experimental_configuration}

Table~\ref{tab:shared_setup} summarizes the experimental settings shared across all three models: DAPF-BERT, BERT-CLS, and BERT-CLS+Prompt. The only differences between models are the input format, prompt components, and prediction interface, which are described in Table~\ref{tab:setup} of the main text.

\begin{table}[t]
\centering
\small
\begin{tabular}{ll}
\toprule
\textbf{Component} & \textbf{Shared setting} \\
\midrule
Backbone & BERT-base-uncased \\
Training data & CCC + ADReSS train \\
Test data & ADReSS test \\
Test size & 48 transcripts \\
Epochs & 10 \\
Batch size & 4 \\
Class weighting & Enabled \\
Random seeds & 0, 1, 2, 3, 4 \\
\bottomrule
\end{tabular}
\caption{Experimental settings shared by DAPF-BERT, BERT-CLS, and BERT-CLS+Prompt.}
\label{tab:shared_setup}
\end{table}

\subsection{Computational Resources}

All experiments were run locally on a MacBook Pro with an Apple M3 Pro chip, a 12-core CPU, an 18-core integrated GPU, and 36 GB of unified memory. PyTorch operations used the Metal Performance Shaders (MPS) backend where GPU acceleration was supported. All three models use BERT-base-uncased, which has approximately 110 million parameters. The complete experimental pipeline, including training across five random seeds, representation probing, Integrated Gradients, Partition SHAP, and deletion and insertion tests, took approximately 2.5 hours on this device.

\section{Partition SHAP Sensitivity Analysis}

As a sensitivity analysis, we applied Partition SHAP to the
DAPF seed whose predictive performance was closest to the
five-seed mean. Following the same evaluation setting used throughout the paper, explanations were generated for the 48 ADReSS test transcripts with respect to the predicted probability of the Dementia class.

Partition SHAP was implemented using SHAP's \texttt{PartitionExplainer} with a maximum evaluation budget of 500 model evaluations per transcript. Unlike the Integrated Gradients analysis in the main paper, this experiment was performed on a single model seed and was intended only to examine whether the qualitative attribution patterns were consistent across attribution methods.

Although the two attribution methods emphasized similar categories of cues, agreement on individual token importance was only moderate. This suggests that the broader qualitative interpretation is relatively stable, whereas precise word-level importance depends on the attribution method. Details can be seen in Table~\ref{tab:shap_compare}.

\begin{table}[h]
\centering
\small
\begin{tabular}{lc}
\toprule
Metric & Value \\
\midrule
IG top-20 tokens with matching SHAP direction & 11 / 20 \\
Spearman correlation & 0.362 \\
\bottomrule
\end{tabular}
\caption{Agreement between Integrated Gradients and Partition SHAP token attributions.}
\label{tab:shap_compare}
\end{table}

To examine the cues emphasized by Partition SHAP, we first applied the same token-reporting and normalization procedure used for Integrated Gradients. Partition SHAP itself was computed on the original, unfiltered prompt-wrapped inputs; filtering was applied only after attribution computation and before token-type aggregation. We excluded prompt-scaffold and special tokens, punctuation-only and control symbols, the artifact \textit{ex}, WordPiece continuations containing one or two characters, and isolated single-letter alphabetic fragments other than \textit{i} and \textit{a}. The remaining tokens were stripped of surrounding whitespace, lowercased, and had WordPiece continuation markers removed, while longer subword fragments were retained.

We then aggregated the resulting attribution scores across the 48 ADReSS test transcripts. For each normalized token type, we calculated its total frequency, the number of transcripts in which it appeared, its mean signed SHAP value, and its mean absolute SHAP value. To reduce the influence of rare tokens, Table~\ref{tab:shap_tokens} includes only tokens that occurred at least 10 times and in at least five transcripts. Positive values indicate that a token contributed, on average, toward the \textsc{Dementia} output, whereas negative values indicate a contribution away from it.

\begin{table*}[t]
\centering
\small
\begin{tabular}{lrrrrlrrrr}
\toprule
\multicolumn{5}{c}{\textbf{Positive Mean SHAP Attribution}} &
\multicolumn{5}{c}{\textbf{Negative Mean SHAP Attribution}} \\
\cmidrule(lr){1-5} \cmidrule(lr){6-10}
Token & Freq. & Trans. & Mean & Mean Abs. &
Token & Freq. & Trans. & Mean & Mean Abs. \\
\midrule
dry   & 12  & 9  & 0.0448 & 0.0464 &
flow      & 23  & 18 & -0.0307 & 0.0349 \\
ing   & 21  & 18 & 0.0428 & 0.0477 &
cookies   & 30  & 21 & -0.0225 & 0.0267 \\
fall  & 22  & 19 & 0.0323 & 0.0335 &
cookie    & 76  & 36 & -0.0222 & 0.0252 \\
reach & 11  & 8  & 0.0278 & 0.0278 &
open      & 15  & 12 & -0.0167 & 0.0172 \\
get   & 23  & 13 & 0.0243 & 0.0284 &
guess     & 10  & 9  & -0.0164 & 0.0218 \\
oh    & 19  & 9  & 0.0176 & 0.0194 &
drying    & 20  & 16 & -0.0155 & 0.0161 \\
well  & 30  & 22 & 0.0162 & 0.0213 &
dish      & 12  & 10 & -0.0151 & 0.0158 \\
yeah  & 12  & 7  & 0.0157 & 0.0160 &
standing  & 20  & 18 & -0.0141 & 0.0161 \\
but   & 20  & 16 & 0.0151 & 0.0169 &
reaching  & 19  & 15 & -0.0132 & 0.0139 \\
got   & 15  & 8  & 0.0119 & 0.0120 &
plate     & 11  & 7  & -0.0114 & 0.0120 \\
tin   & 14  & 8  & 0.0109 & 0.0109 &
from      & 14  & 10 & -0.0114 & 0.0137 \\
and   & 240 & 45 & 0.0104 & 0.0138 &
think     & 12  & 11 & -0.0105 & 0.0158 \\
some  & 20  & 10 & 0.0096 & 0.0105 &
getting   & 12  & 11 & -0.0097 & 0.0117 \\
gonna & 14  & 10 & 0.0086 & 0.0162 &
is        & 169 & 41 & -0.0092 & 0.0132 \\
him   & 15  & 12 & 0.0084 & 0.0102 &
stool     & 61  & 40 & -0.0083 & 0.0134 \\
\bottomrule
\end{tabular}
\caption{Tokens with the highest positive and negative mean Partition SHAP attributions across the ADReSS test set after applying the same token-reporting and normalization procedure used for Integrated Gradients. Only tokens occurring at least 10 times and in at least five transcripts are included. \textit{Freq.} is the total number of occurrences, \textit{Trans.} is the number of transcripts containing the token, \textit{Mean} is the mean signed attribution to the \textsc{Dementia} output, and \textit{Mean Abs.} is the mean absolute attribution.}
\label{tab:shap_tokens}
\end{table*}

The positive and negative rankings contain a mixture of picture-description vocabulary, discourse forms, and tokenization fragments. Words such as \textit{dry}, \textit{fall}, \textit{reach}, \textit{cookie}, \textit{drying}, and \textit{standing} are directly related to the Cookie Theft task. Discourse forms such as \textit{oh}, \textit{well}, \textit{yeah}, and \textit{guess} also receive relatively large attributions, while longer fragments such as \textit{ing} and \textit{tin} suggest sensitivity to tokenization conventions. Semantically related forms do not always receive the same direction: for example, \textit{dry} has a positive mean attribution, whereas \textit{drying} has a negative one; similarly, \textit{reach} is positive while \textit{reaching} is negative. These differences suggest that individual token directions should not be treated as stable linguistic evidence. Overall, Partition SHAP identifies the same broad types of cues seen with Integrated Gradients, but the two methods do not consistently agree on which individual tokens are most important.

To conclude, Partition SHAP highlighted the same broad categories of evidence identified by Integrated Gradients, including picture-description vocabulary, discourse markers, and transcription artifacts. However, individual words often received different attribution magnitudes or even opposite attribution directions, indicating that token-level explanations remain sensitive to the choice of attribution method.

\section{Faithfulness Details}
\label{app:faithfulness}
Table~\ref{tab:faith-positive-compare-sd} extends Table~\ref{tab:faith-positive-compare} by reporting standard deviations across random seeds.

\begin{table}[t]
\centering
\small
\renewcommand{\arraystretch}{1.1}
\begin{tabular}{L{1.45cm}L{1.1cm}C{1cm}C{1cm}C{1cm}}
\toprule
\textbf{Model} & \textbf{Test} & \textbf{5\%} & \textbf{20\%} & \textbf{50\%} \\
\midrule
DAPF & Deletion Drop & 0.0080 \scriptsize{$\pm$ 0.0155} & 0.0055 \scriptsize{$\pm$ 0.0132} & -0.0318 \scriptsize{$\pm$ 0.0144} \\
DAPF & Insertion Recover & -0.0187 \scriptsize{$\pm$ 0.0192} & -0.0352 \scriptsize{$\pm$ 0.0395} & -0.0916 \scriptsize{$\pm$ 0.0612} \\
\midrule
BERT-CLS & Deletion Drop & 0.0259 \scriptsize{$\pm$ 0.0287} & 0.0142 \scriptsize{$\pm$ 0.0331} & 0.0067 \scriptsize{$\pm$ 0.0280} \\
BERT-CLS & Insertion Recover & 0.0369 \scriptsize{$\pm$ 0.0226} & 0.0813 \scriptsize{$\pm$ 0.0392} & 0.1223 \scriptsize{$\pm$ 0.0558} \\
\midrule
BERT-CLS +Prompt & Deletion Drop & 0.0302 \scriptsize{$\pm$ 0.0277} & 0.0426 \scriptsize{$\pm$ 0.0409} & 0.0259 \scriptsize{$\pm$ 0.0444} \\
BERT-CLS +Prompt & Insertion Recover & 0.0279 \scriptsize{$\pm$ 0.0185} & 0.0787 \scriptsize{$\pm$ 0.0291} & 0.1225 \scriptsize{$\pm$ 0.0417} \\
\bottomrule
\end{tabular}
\caption{Positive-ranking faithfulness comparison using tokens ranked by positive attribution toward \textsc{Dementia}. Positive values indicate behavior consistent with dementia-supporting evidence. Results are shown for 5\%, 20\%, and 50\% perturbation fractions.}
\label{tab:faith-positive-compare-sd}
\end{table}

Table~\ref{tab:faith-absolute-compare} reports faithfulness using tokens ranked by absolute attribution magnitude. Negative probability drops can thus occur when the highest-magnitude tokens mostly push the model away from \textsc{Dementia}, and we observe this pattern for all three models. 
DAPF shows negative deletion and insertion effects overall. 
BERT-CLS shows increasingly negative deletion effects, and its insertion effect becomes negative at 50\%. 
BERT-CLS+Prompt shows the strongest negative deletion effects, indicating that many high-magnitude tokens provide evidence against \textsc{Dementia}.

\begin{table}[t]
\centering
\small
\renewcommand{\arraystretch}{1.1}
\begin{tabular}{L{1.45cm}L{1.1cm}C{1cm}C{1cm}C{1cm}}
\toprule
\textbf{Model} & \textbf{Test} & \textbf{5\%} & \textbf{20\%} & \textbf{50\%} \\
\midrule
DAPF & Deletion Drop & -0.0511 \scriptsize{$\pm$ 0.0471} & -0.0939 \scriptsize{$\pm$ 0.0499} & -0.1799 \scriptsize{$\pm$ 0.0483} \\
DAPF & Insertion Recover & -0.0709 \scriptsize{$\pm$ 0.0607} & -0.1902 \scriptsize{$\pm$ 0.1175} & -0.3453 \scriptsize{$\pm$ 0.1019} \\
\midrule
BERT-CLS & Deletion Drop & -0.0166 \scriptsize{$\pm$ 0.0267} & -0.1318 \scriptsize{$\pm$ 0.0524} & -0.3318 \scriptsize{$\pm$ 0.1184} \\
BERT-CLS & Insertion Recover & 0.0407 \scriptsize{$\pm$ 0.0247} & 0.0082 \scriptsize{$\pm$ 0.0353} & -0.1940 \scriptsize{$\pm$ 0.1054} \\
\midrule
BERT-CLS+Prompt & Deletion Drop & -0.0997 \scriptsize{$\pm$ 0.0775} & -0.2207 \scriptsize{$\pm$ 0.0907} & -0.3883 \scriptsize{$\pm$ 0.1016} \\
BERT-CLS+Prompt & Insertion Recover & 0.0008 \scriptsize{$\pm$ 0.0102} & -0.0189 \scriptsize{$\pm$ 0.0395} & -0.2402 \scriptsize{$\pm$ 0.0823} \\
\bottomrule
\end{tabular}
\caption{Absolute-ranking faithfulness. Results should not be interpreted as target-specific rationales.}
\label{tab:faith-absolute-compare}
\end{table}

\end{document}